\documentclass[sigconf]{acmart}

\usepackage{CJKutf8}
\usepackage{caption}
\usepackage{amsmath,amsfonts}
\usepackage{algorithmic}
\usepackage{array}
\usepackage{algorithm}
\usepackage{multirow}
\usepackage{bbding}
\usepackage{array}
\usepackage{makecell}
\usepackage{amsmath}
\usepackage{xspace}
\usepackage{color}

\usepackage{amssymb}
\usepackage{booktabs}
\usepackage{pifont}
\usepackage{bm}
\usepackage{soul}
\usepackage{hyperref}
\hypersetup{
    colorlinks=true, 
    urlcolor=magenta,   
}

\AtBeginDocument{%
  }

\copyrightyear{2025} 
\acmYear{2025} 
\setcopyright{acmlicensed}\acmConference[MM '25]{Proceedings of the 33rd
ACM International Conference on Multimedia}{October 27--31, 2025}{Dublin,
Ireland}
\acmBooktitle{Proceedings of the 33rd ACM International Conference on
Multimedia (MM '25), October 27--31, 2025, Dublin, Ireland}
\acmDOI{10.1145/3746027.3755036}
\acmISBN{979-8-4007-2035-2/2025/10}

\begin{document}
\begin{CJK}{UTF8}{gbsn}
\title{Learning from Heterogeneity: Generalizing Dynamic Facial Expression Recognition via Distributionally Robust Optimization}

\def\eg{\textit{e.g.}}
\def\ie{\textit{i.e.}}
\def\etal{\textit{et al.}}
\def\etc{\textit{etc.}}
\def\whp{\textit{w.h.p.}}
\def\iid{\textit{i.i.d.}}
\def\name{\emph{HDF}}
\def\MA{\emph{DAM}}
\def\MB{\emph{DSM}}


\author{Feng-Qi Cui}
\authornote{Both authors contributed equally to this research.}

\orcid{0000-0002-0454-940X}
\affiliation{%
  \institution{University of Science and \\Technology of China}
   \city{Hefei}
  \country{China}
  }
\email{fengqi_cui@mail.ustc.edu.cn}

\author{Anyang Tong}
\authornotemark[1]
\orcid{0000-0003-0960-1497}
\affiliation{%
  \institution{Hefei University of Technology}
  \city{Hefei}
  \country{China}
  }
  \email{tonganyang@mail.hfut.edu.cn}

\author{Jinyang Huang}
\authornote{Corresponding Author. E-mail: hjy@hfut.edu.cn}
\orcid{0000-0001-5483-2812}
\affiliation{%
  \institution{Hefei University of Technology}
  \city{Hefei}
  \country{China}
  }
  \email{hjy@hfut.edu.cn}
  
\author{Jie Zhang}
\orcid{0000-0002-4230-1077}
\affiliation{%
    \institution{IHPC and CFAR, A*STAR}
  \city{Singapore}
   \country{Singapore}
  }
  \email{zhang_jie@cfar.a-star.edu.sg}
  
\author{Dan Guo}
\orcid{0000-0003-2594-254X}

\affiliation{%
  \institution{Hefei University of Technology}
   \city{Hefei}
   \country{China}
  }
  \email{guodan@hfut.edu.cn}

\author{Zhi Liu}
\orcid{0009-0004-2035-8373}
\affiliation{%
  \institution{The University of Electro-Communications}
  \city{Tokyo}
  \country{Japan}
  }
  \email{liuzhi@uec.ac.jp}

\author{Meng Wang}
\orcid{0000-0002-3094-7735}
\affiliation{%
  \institution{Hefei University of Technology}
   \city{Hefei}
  \country{China}
  }
  \email{eric.mengwang@gmail.com}







\begin{abstract}

Dynamic Facial Expression Recognition (DFER) plays a critical role in affective computing and human-computer interaction. Although existing methods achieve comparable performance, they inevitably suffer from performance degradation under sample heterogeneity caused by multi-source data and individual expression variability.
To address these challenges, we propose a novel framework, called \textbf{Heterogeneity-aware Distributional Framework} (\name), and design two plug-and-play modules to enhance time-frequency modeling and mitigate optimization imbalance caused by hard samples. 
Specifically, the \textbf{Time-Frequency Distributional Attention Module} (\MA) captures both temporal consistency and frequency robustness through a dual-branch attention design, improving tolerance to sequence inconsistency and visual style shifts. 
Then, based on gradient sensitivity and information bottleneck principles, an adaptive optimization module \textbf{Distribution-aware Scaling Module} (\MB) is introduced to dynamically balance classification and contrastive losses, enabling more stable and discriminative representation learning. 
Extensive experiments on two widely used datasets, DFEW and FERV39k, demonstrate that \name~significantly improves both recognition accuracy and robustness. Our method achieves superior weighted average recall (WAR) and unweighted average recall (UAR) while maintaining strong generalization across diverse and imbalanced scenarios. Codes are released at \url{https://github.com/QIcita/HDF_DFER}.
\end{abstract}
\maketitle

\begin{figure}[t]
\centering
\includegraphics[width=0.48\textwidth]{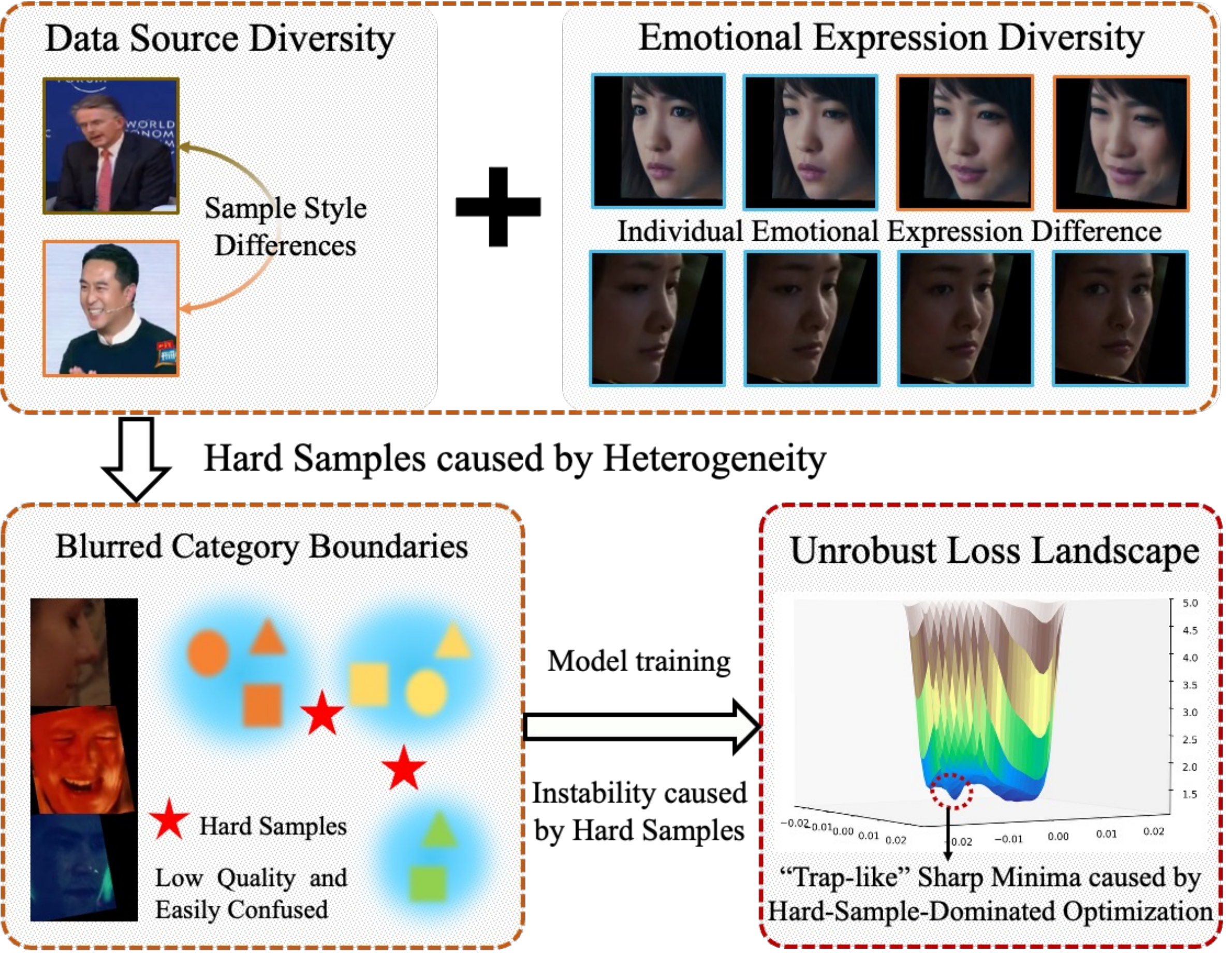} 
\caption{ Due to a series of Sample Heterogeneity problems, there are a large number of Hard Samples in DFER, which leads to fuzzy category boundaries and unstable loss optimization process.}
\vspace{-0.3cm}
\label{intro}
\end{figure}

\section{Introduction}
Facial expression recognition (FER) can automatically perceive human emotions and has been widely applied in real-world scenarios such as psychological evaluation \cite{Uddin_Joolee_Lee_2022, guo2024benchmarking,10149418}, and human-computer interaction \cite{10.1145/3581783.3612342,Melo_Granger_Hadid_2022,guo2018hierarchical}.
According to the data type, FER approaches can be divided into two categories: Static FER (SFER) \cite{10.1145/3581783.3611702} and Dynamic FER (DFER) \cite{Zhang_Tian_Zhang_Guo_Xu}. SFER methods focus on analyzing single still images to classify emotions \cite{9585378}, which limits their capacity to capture the dynamic progression and contextual transitions of expressions across time \cite{li2020deep}, leading to degraded performance in realistic settings that involve temporal complexity and motion variance. In contrast, DFER techniques utilize video sequences to model how facial expressions evolve over time, offering a more faithful representation of emotional dynamics. However, it is challenging to understand the spatiotemporal expressions in real scenarios.

Considering that facial expressions are usually subtle, transitory, and subject to spatio-temporal variations in the real world, pioneer works have been done to capture the subtle changes in expressions through temporal modeling. For example, recent DFER methods aim to model the temporal evolution and structural dynamics of facial expressions by integrating spatiotemporal cues through CNN-RNN pipelines \cite{wang2019multi, dfew}, unified 3D CNNs like I3D \cite{i3d} and M3DFEL \cite{m3dfel}, or transformer-based attention mechanisms as in Former-DFER \cite{formerDFER} and STT \cite{sttMa_Sun_Li_2022}. Although these temporal modeling capabilities enhance the robustness and adaptability of DFER in real-world environments and further achieve impressive performances \cite{wang2024facialpulse}, several problems still need to be seriously addressed before widespread application.

As illustrated in Fig.~\ref{intro}, due to the lack of mechanisms to handle hard samples arising from real-world sample heterogeneity, which is extremely common and harmful, the biggest problem with existing DFER methods is poor generalization. The heterogeneity basically caused by data source differences, individual differences, and training dynamics. 
\emph{1) Overlooking source-level heterogeneity undermines robustness to visual variation.} DFER datasets are typically built from heterogeneous video sources such as movies, news, and interviews, which introduce substantial variations in visual style, resolution, compression, and motion continuity. These source-level discrepancies manifest as sample-level distributional variations, which make it difficult to learn consistent and discriminative representations.
Without explicitly modeling the heterogeneity in visual styles and quality, models tend to overfit dominant patterns and struggle with underrepresented or atypical samples, ultimately degrading generalization under real-world conditions.
\emph{2) Individual-induced heterogeneity leads to hard samples and representation inconsistency.} Variations in identity, expression patterns, age, and emotional intensity lead to significant individual-induced heterogeneity, where the same emotion is expressed in diverse and often inconsistent ways. This results in increased intra-category variation and a higher prevalence of hard or ambiguous samples during training, which undermines representation stability and degrades the reliability of learned decision boundaries.
\emph{3) Overlooking training dynamics under sample heterogeneity hinders optimization stability.} In the presence of sample heterogeneity, data-driven deep models often encounter optimization difficulties, such as fluctuating gradients and unstable convergence. Hard samples dominate training signals and may overwhelm classification loss, further leading to brittle representations. 
 
To address the above challenges in DFER, inspired by the principles of Distributionally Robust Optimization, we propose a novel framework named \textbf{Heterogeneity-aware Distributional Framework} (\name), which incorporates two plug-and-play DFER-enhancing modules that can be seamlessly integrated into existing general-purpose video backbone networks. First, to construct more stable and generalizable expression representations under sample heterogeneity, we design the \textbf{Time-Frequency Distributional Attention Module} (\MA). By leveraging wasserstein-regularized temporal attention and frequency-domain adversarial perturbation to capture highly uncertain expression dynamics and suppress distributional variations in style and resolution, \MA~adopts a dual-branch architecture to perform adaptive and robust modeling in both temporal and frequency domains.
To further mitigate the optimization instability and representation degradation caused by hard samples, we introduce a novel \textbf{Distribution-aware Scaling Module} (\MB). By combining an information-constrained and reweighted supervised contrastive loss with a sharpness-aware loss scaling mechanism driven by gradient norm, \MB~constructs a dual-enhanced optimization strategy that adaptively balances classification and contrastive learning objectives, thereby improving both the robustness and discriminability of learned representations under heterogeneous conditions.

Totally, our main contributions can be concluded as follows:

\begin{itemize}
\item To the best of our knowledge, this paper is the first attempt to introduce the principle of Distributionally Robust Optimization into DFER. By joint modeling of style variations, individual expression dynamics, and task-level distribution shifts across both feature and optimization levels, we construct \name~that explicitly addresses ubiquitous sample heterogeneity.

\item A novel distributionally robust attention module \MA~is first proposed by adopting a dual-branch structure to jointly model temporal consistency and frequency robustness under sample heterogeneity. The frequency branch integrates adversarial perturbation, and dynamic activation to improve generalization across diverse visual styles, while the temporal branch leverages wasserstein-regularized attention to stabilize expression evolution across individuals.

\item We introduce a new optimization strategy named \MB, which combines distributionally robust contrastive learning with gradient-inspired adaptive loss scaling mechanism. This dual design enables dynamic balancing between classification supervision and representation learning, improving training stability and generalization under multi-source and heterogeneous sample scenarios.

\item Extensive experiments conducted on two challenging in-the-wild DFER datasets, DFEW and FERV39k, demonstrate the effectiveness of our approach. The proposed \name~ significantly outperforms state-of-the-art methods in recognition accuracy, and robustness to sample heterogeneity.

\end{itemize}
\section{Related work}
\subsection{Dynamic Facial Expression Recognition}
Dynamic Facial Expression Recognition (DFER) focuses on modeling temporal changes in facial expressions, capturing subtle emotional dynamics beyond static images. Early methods relied on hand-crafted spatiotemporal features such as LBP-TOP \cite{4160945} and HOG-TOP \cite{7518582}. However, the property of dependence on prior knowledge limited generalization. With the rise of deep learning, CNN-RNN frameworks became mainstream, which using CNNs ( \eg, VGG \cite{8275511}, ResNet \cite{resnHe_2016_CVPR}) for spatial encoding and LSTM/GRU \cite{9102419} for temporal modeling. Unfortunately, such methods often struggle to capture long-range dependencies and are sensitive to distribution shifts, limiting their robustness in real-world scenarios.
To jointly learn spatial and temporal features, 3D CNNs such as C3D \cite{c3d}, I3D \cite{i3d}, and M3DFEL \cite{m3dfel} have been widely used. M3DFEL incorporates multi-instance learning to suppress noisy frames, while IAL \cite{ial/aaai.v37i1.25077} introduces intensity loss to enhance inter-category separability. However, these models often struggle with ambiguous boundaries and sample uncertainty. 

To improve representation quality, state-of-the-art methods explore vision-language modeling \cite{10687508,10.1145/3664647.3681583,10.1145/3664647.3680827}, label-guided temporal fusion \cite{Chen_Wen_Yang_Li_Chen_Wang}, or static-to-dynamic domain adaptation \cite{Zhang_Tian_Zhang_Guo_Xu, huang2023phyfinatt, hu2025unified}, aiming to leverage additional supervision. Yet, they are still constrained by reliance on large-scale pre-trained data and limited temporal reasoning capability. Transformer-based architectures have gained popularity due to their ability to capture long-range dependencies through self-attention mechanisms. For instance, former-DFER \cite{formerDFER} and STT \cite{sttMa_Sun_Li_2022} integrate spatiotemporal attention to improve robustness against occlusion and motion variation. Nevertheless, these transformer-based DFER models inevitably face challenges in effectively handling sample heterogeneity, noisy transitions, and data distribution shifts, further limiting their applicability to complex real-world scenarios.

\subsection{Distributionally Robust Optimization}
Distributionally Robust Optimization provides a principled framework for improving generalization under sample heterogeneity \cite{wu2023understanding}. Unlike conventional robust optimization that addresses local input perturbations \cite{jin2020sampling,10122715}, Distributionally Robust Optimization minimizes the worst-case expected loss over a family of plausible data distributions, thereby accounting for global distribution shifts. These uncertainty sets are commonly defined using $\phi$-divergence \cite{duchi2019variance}, Wasserstein distance \cite{sinha2018certifiable, huang2022coresets}, or structural constraints \cite{10.1287/opre.2020.1990}, which enable the model to handle latent distributional variations beyond observed data noise.
Although recent work has extended Distributionally Robust Optimization with learnable uncertainty sets for cross-domain robustness \cite{michel2022distributionally}, directly applying Distributionally Robust Optimization in high-dimensional and temporally-evolving tasks such as DFER still remains challenging.

Thus, in this work, we explore Distributionally Robust Optimization as a principled solution to cope with sample heterogeneity in DFER, which arises from variations in visual style, expression dynamics, and individual identity, \etc~To address this, we propose heterogeneity-aware Distributionally Robust Optimization framework \name~tailored for DFER. Among \name, \MA~models sample heterogeneity in temporal dynamics and visual styles via dual-branch attention, while \MB~stabilizes training under hard samples by adaptively balancing classification and representation objectives based on gradient sensitivity. Furthermore, these components form a unified heterogeneity-aware framework guided by Distributionally Robust Optimization, enabling robust learning from diverse and uncertain dynamic expression samples.
\section{Preliminaries}

\subsection{Temporal-Frequency Representation}

Temporal and frequency representations provide two complementary perspectives for understanding dynamic facial expressions. The temporal domain focuses on the evolution of facial expressions over time, highlighting how expressions change across consecutive frames. Given a video sequence \( \{x_1, x_2, \dots, x_t\} \), temporal differences between adjacent frames can be modeled as:
\begin{equation}
\Delta x_t = x_t - x_{t-1},
\end{equation}
which captures frame-wise motion cues and short-term expression dynamics.
In parallel, frequency-domain analysis encodes facial appearance using spatial frequency components, providing a robust view of structural texture and detail variations. The Discrete Cosine Transform (DCT) \cite{9710319, huang2024keystrokesniffer, ZHU2020116}, commonly used for frequency feature extraction, is applied to each frame $x \in \mathbb{R}^{H \times W}$ to obtain a frequency representation $X_f \in \mathbb{R}^{H \times W}$ defined as:
\begin{equation}
X_f(u, v) = \alpha(u)\alpha(v) \sum_{i,j} x(i, j) \cdot 
\cos\left[\frac{\pi (2i+1)u}{2H}\right] 
\cos\left[\frac{\pi (2j+1)v}{2W}\right],
\end{equation}
with normalization terms:
\begin{equation}
\alpha(u) = 
\begin{cases}
\sqrt{\frac{1}{H}}, & \text{if } u = 0 \\
\sqrt{\frac{2}{H}}, & \text{otherwise}
\end{cases}, \quad
\alpha(v) = 
\begin{cases}
\sqrt{\frac{1}{W}}, & \text{if } v = 0 \\
\sqrt{\frac{2}{W}}, & \text{otherwise}
\end{cases}.
\end{equation}

These two representations can be integrated to form a unified feature embedding that incorporates both temporal motion and frequency-based appearance information:
\begin{equation}
z = \lambda_t \cdot z_t + \lambda_f \cdot z_f,
\end{equation}
where \( z_t \) and \( z_f \) are temporal and frequency features respectively, and \( \lambda_t, \lambda_f \in [0,1] \) are learnable weights.
This formulation provides the theoretical foundation for the time-frequency modeling in our method, enabling downstream modules to capture both motion dynamics and structural variations essential for robust dynamic facial expression representation.

\subsection{Supervised Contrastive Loss}

Contrastive learning (CL) \cite{zhou2022contrastive} aims to learn discriminative representations by pulling semantically similar samples closer and pushing dissimilar ones apart in the embedding space. While traditional CL is commonly unsupervised and constructs positive pairs via data augmentation, supervised contrastive learning (SCL) leverages label information to form more informative positive and negative pairs. 
Given a batch of $N$ samples, let $\mathcal{I}=\{1,\dots,N\}$ denote the index set. For an anchor sample $i \in \mathcal{I}$, let $P(i)$ be the set of indices (excluding $i$) that share the same class label with $i$. The SCL is defined as:
\begin{equation}
\mathcal{L}_{\text{SC}} = \sum_{i \in \mathcal{I}} \frac{1}{|P(i)|} \sum_{p \in P(i)} -\log \frac{\exp(\mathbf{z}_i \cdot \mathbf{z}_p / \tau)}{\sum_{a \in \mathcal{A}(i)} \exp(\mathbf{z}_i \cdot \mathbf{z}_a / \tau)}
\end{equation}
where $\mathbf{z}_i \in \mathbb{R}^d$ is the normalized representation of sample $i$, $\tau$ is a temperature scalar, and $\mathcal{A}(i)$ is the set of all contrastive candidates.
SCL encourages compact intra-category clusters in the embedding space, reducing the need for hard sample mining and enhancing both classification and generalization.

\section{Methodology}
 
\begin{figure*}[t]
\centering
\includegraphics[width=0.93\textwidth]{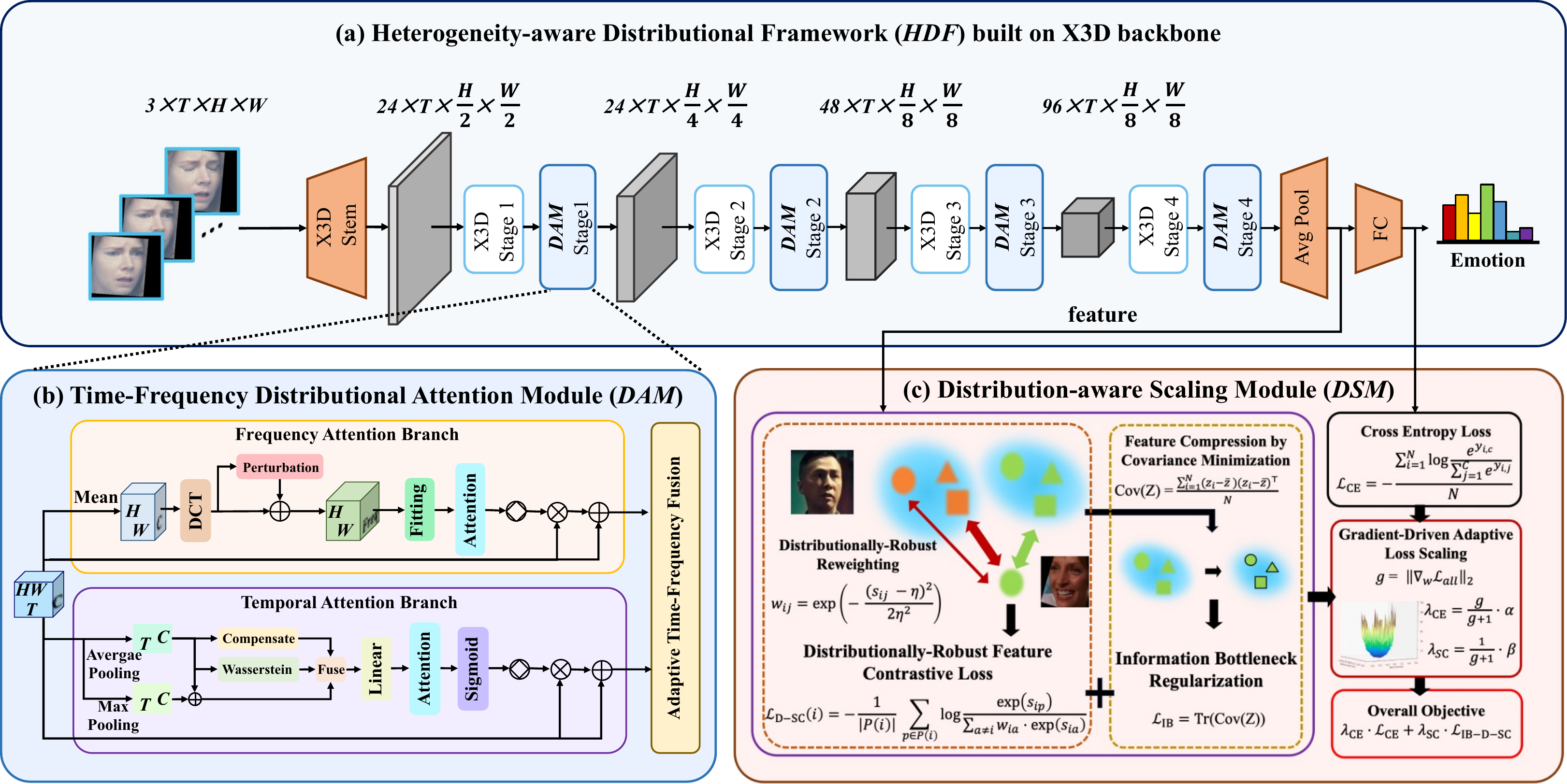} 
\caption{An overview of the proposed \name. (a) The overall processing flow of \name. (b) The pipeline of The pipeline of Time-Frequency Distributional Attention (\MA) in \name. (c) The sketch of Distribution-aware Scaling (\MB).}
\vspace{-0.4cm}
\label{ours}
\end{figure*}

We adopt X3D \cite{Feichtenhofer_2020_CVPR} as the feature extraction backbone and embed the two plug-and-play modules (the Time-Frequency Distributional Attention Module (\MA) in Sec. 4.2 and the Distribution-aware Scaling Module (\MB) in Sec. 4.3) we designed. Specifically, \MA~enhances feature robustness under sample heterogeneity by modeling temporal consistency and frequency stability. \MB~adaptively balances classification and contrastive losses based on gradient sharpness, enabling more stable and discriminative optimization. The workflow of the proposed \name~is illustrated in Fig. \ref{ours}. 

\subsection{Distributionally-Robust Attention on Frequency and Time}
 
We propose Time-Frequency Distributional Attention Module (\MA), a dual-branch attention architecture designed to jointly enhance temporal consistency and style robustness in dynamic facial expression representation. 
Sample heterogeneity driven by source diversity and individual differences disrupts stable training and degrades feature consistency, which are insufficiently handled by conventional empirical risk minimization.
To address this issue, \MA~incorporates a temporal attention branch and a frequency attention branch, both guided by Distributionally Robust Optimization principles. These branches are responsible for modeling temporal inconsistency and visual heterogeneity, respectively. A learnable fusion mechanism further adaptively integrates the two modalities, enhancing feature stability and generalization under diverse and heterogeneous expression conditions.

\subsubsection{Frequency Attention Branch: Distributionally Robust Modeling via Adversarial Perturbation and Dynamic Fitting}

To address sample heterogeneity caused by variations in visual style, image quality, and illumination, \MA~introduces a frequency attention branch, which integrates DCT-based frequency decomposition, distribution-aware adversarial perturbation, and dynamic activation adjustment to improve robustness against source-specific visual shifts.

First, the input feature x $\in \mathbb{R}^{B \times T \times C \times H \times W}$ is passed through a DCT convolution to extract frequency features $f_{\text{DCT}}$.
To simulate worst-case variations caused by source-level sample heterogeneity and enhance robustness to visual style shifts, adversarial perturbation introduce two types of regularized perturbations to the original frequency features. 
The first is a sign-based perturbation: $\alpha \cdot \epsilon \cdot \text{sign}(f_{\text{DCT}})$, where $\text{sign}(\cdot)$ denotes the element-wise sign function, generating adversarial noise along the gradient direction; $\alpha$ is a learnable weight and $\epsilon$ is a fixed scaling factor. The second is a wasserstein-guided global shift: $\beta \cdot \text{sign}(f_{\text{DCT}}) \cdot \|f_{\text{DCT}} - \mathbb{E}[f_{\text{DCT}}]\|2$, where $\mathbb{E}[f_{\text{DCT}}]$ denotes the mini-batch mean of frequency features, reflecting the global distribution center; $\|\cdot\|_2 $is the L2 norm, measuring the deviation between a sample and the global distribution; and $\beta$ is a learnable global perturbation weight.
The resulting adversarial frequency feature is:
\begin{equation}
f_{\text{adv}} = f_{\text{DCT}} + \alpha \cdot \epsilon \cdot \text{sign}(f_{\text{DCT}}) + \beta \cdot \text{sign}(f_{\text{DCT}}) \cdot \left\| f_{\text{DCT}} - \mathbb{E}[f_{\text{DCT}}] \right\|_2 .
\end{equation}
This design approximates the inner maximization problem in Distributionally Robust Optimization by jointly simulating input-level perturbations and feature-level distributional deviations, enabling the model to better handle frequency-domain variations caused by source-specific heterogeneity.
To address activation instability from perturbations, we design a dynamic fitting function that adaptively scales activations:
\begin{equation}
f_{\text{dyn}} = \tanh\left( \frac{\alpha}{\sqrt{\text{Var}(f_{\text{adv}}) + \varepsilon}} \cdot f_{\text{adv}} \right) \cdot \gamma + \beta,
\end{equation}
where $\alpha$ is a learnable scaling factor, $\gamma$ and $\beta$ are affine transformation parameters, and $\varepsilon$ is a small constant for numerical stability. $\sqrt{\text{Var}(f_{\text{adv}})}$ denotes the standard deviation for each sample, which reflects the dispersion of activation values. This function dynamically adjusts activation strength according to channel-wise feature statistics, mitigating inter-sample scale variation and improving robustness under visual heterogeneity.

Finally, the dynamically activated feature $f_{\text{dyn}}$ is processed by self-attention mechanism to obtain the attention map $f_{\text{att}}$, which is then fused with the original features:
\begin{equation}
x_s = x \cdot \sigma(f_{\text{att}}) + x,
\end{equation}
where $\sigma(\cdot)$ denotes the sigmoid function. In summary, the frequency branch improves feature robustness against source-induced style variation by jointly leveraging frequency decomposition, adversarial modeling, and adaptive activation fitting.

\subsubsection{Temporal Attention Branch: Sequence Consistency Modeling via Wasserstein Regularization}

In DFER, individual expression differences exacerbate temporal inconsistencies in frame-level features, challenging consistent trajectory modeling. To mitigate this temporal heterogeneity, we introduce a temporal attention branch that enhances robustness against dynamic inconsistency and irregular expression rhythms.

Given the input feature $x$, we compute two temporal descriptors for each frame by applying global average pooling and max pooling over the channel and spatial dimensions, denoted as $A_{\text{avg}}^{(t)}$ (global context) and $A_{\text{max}}^{(t)}$ (local saliency), respectively.
To measure the deviation of the current frame from the global expression trajectory, we introduce a wasserstein-inspired regularization term, approximated as the L2 distance between the average representation of the current frame and the global mean of the sequence: $\mathcal{W}_{\text{global}}^{(t)} = \left\| A_{\text{avg}}^{(t)} - \mathbb{E}_{t'}[A_{\text{avg}}^{(t')}] \right\|_2$, which is used to identify outlier or drifting frames that compromise the temporal consistency of the expression sequence. To prevent this wasserstein term from suppressing genuine emotional transitions, we further introduce a local temporal difference term as a compensatory measure: $\mathcal{D}_{\text{local}}^{(t)} = \left\| A_{\text{avg}}^{(t)} - A_{\text{avg}}^{(t-1)} \right\|_2, $
which enhances the model’s sensitivity to short-term dynamic variations. We then fuse the above components using a gating mechanism to generate the final temporal attention map:
\begin{equation}
\text{TA}_{t,t'} = \sigma \left( \gamma (\alpha \tilde{A}_{\text{avg}} + \beta \tilde{A}_{\text{max}}) + \delta \cdot \mathcal{W}_{\text{global}} - \mathcal{D}_{\text{local}} \right),
\end{equation}
where $\alpha, \beta, \gamma, \delta$ are learnable fusion weights and $\sigma(\cdot)$ denotes the sigmoid activation function. This attention map adaptively adjusts the contribution of each frame to the overall sequence representation. 
Finally, $\text{TA}_{t,t'}$ are passed through a self-attention mechanism to generate a temporal attention map $\text{TA}$, which adaptively highlights consistent frames and suppresses noisy or drifting ones. The attended sequence representation is then fused with the original temporal features as:
\begin{equation}
x_t =  x \cdot \sigma(\text{TA}) + x,
\end{equation}
where $\sigma(\cdot)$ denotes the sigmoid function.
In summary, the temporal branch improves sequence-level consistency under heterogeneous conditions by integrating wasserstein-based global deviation detection, local dynamic compensation, and long-range temporal attention, providing robust and coherent temporal representations for the subsequent fusion module.

\subsubsection{Adaptive Time-Frequency Fusion: Robust Integration of Heterogeneous Modalities}
To unify the complementary strengths of the temporal and frequency branches under sample heterogeneity, we introduce an adaptive time-frequency fusion mechanism that dynamically adjusts the contribution of each modality based on input-specific characteristics. Concretely, we apply global average pooling to the outputs of the temporal branch $x_t$ and the frequency branch $x_s$, obtaining compact descriptors $\bar{x}_t and \bar{x}_s$, respectively. These descriptors are fed into learnable gating functions to compute soft fusion weights:
\begin{equation}
\lambda_t = \sigma(W_t \bar{x}_t + b_t), \quad \lambda_s = \sigma(W_s \bar{x}_s + b_s),
\end{equation}
with a normalization constraint $\lambda_t + \lambda_s = 1$. The fused feature representation is then computed as:
\begin{equation}
x_{\text{fused}} = \lambda_t \cdot x_t + \lambda_s \cdot x_s.
\end{equation}

This fusion mechanism adaptively integrates temporal and frequency information based on input dynamics, allowing the model to attend more to temporal cues under sequence inconsistency and to frequency cues under style-level variations. The resulting representation is thus more robust to sample heterogeneity and better suited for expression classification.

\subsection{Distribution-Aware Robust Scaling for Dynamic Expression Learning}

We propose a novel optimization module called Distribution-aware Scaling Module (\MB), which aims to enhance model robustness and generalization in multi-source heterogeneous data scenarios. \MB~integrates two complementary components: a supervised contrastive loss (SCL) with information bottleneck regularization and distributionally robust reweighting, and a gradient-driven adaptive loss scaling strategy inspired by Sharpness-Aware Minimization (SAM) \cite{foret2021sharpnessaware}. These modules work in tandem to enable the model to learn compact and discriminative representations, while dynamically adjusting the objective weights according to the optimization state throughout training.

\subsubsection{Information-Constrained Distributionally-Robust Contrastive Loss.}
In DFER, hard samples caused by sample heterogeneity not only increase feature distribution diversity but also significantly undermine the stability of positive-negative discrimination in contrastive learning. To address this, we propose an information bottleneck and Distributionally Robust Optimization enhanced SCL that effectively mitigates the instability caused by hard samples during training.
First, to reduce the dominance of hard negative samples in the contrastive loss, we introduce a gaussian kernel-based reweighting strategy into SCL, assigning distributionally robust weights to each negative pair based on their similarity. This design suppresses the risk of overfitting to extreme samples at the distribution level. Let the normalized feature representations be $z_i, z_j \in \mathbb{R}^d$, and the similarity between sample $i$ and $j$ is defined as: $s_{ij} = \frac{z_i^\top z_j}{\tau}$, where $\tau$ is a temperature parameter. For each negative sample pair $(i, j)$, we construct a gaussian robust weight term based on the deviation of their similarity from the $\eta$: $ w_{ij} = \exp\left( - \frac{(s_{ij} - \eta)^2}{2\eta^2} \right)$, where $\eta$ is a hyperparameter that defines the center and smoothness of the gaussian weighting curve, reflecting the expected similarity between well-aligned features, and adjusts how strongly outlier negatives are down weighted. The resulting distributionally-robust SCL is defined as:
\begin{equation}
\mathcal{L}_{\text{D-SC}}^{(i)} = - \frac{1}{|P(i)|} \sum{p \in P(i)} \log \frac{\exp(s_{ip})}{\sum_{a \neq i} w_{ia} \cdot \exp(s_{ia})},
\end{equation}
where $P(i)$ denotes the set of positive samples of anchor $i$. This design significantly reduces the contribution of alignment errors with hard negatives in the loss function, thereby enhancing the stability of the optimization.

On the other hand, we further incorporate the information bottleneck principle by minimizing the mutual information between the representation and the input, aiming to compress redundant information and enhance the task relevance and discriminability of the learned features. To incorporate the information bottleneck constraint, we define the feature covariance matrix as:
\begin{equation}
\text{Cov}(Z) = \frac{1}{N} \sum_{i=1}^{N} (z_i - \bar{z}) (z_i - \bar{z})^\top,
\end{equation}
and approximate the information bottleneck regularization term as: $\mathcal{L}_{\text{IB}} = \text{Tr}(\text{Cov}(Z))$. The final IB-Distributionally Robust Optimization SCL is:
\begin{equation}
\mathcal{L}_{\text{IB-D-SC}} = \frac{1}{N} \sum_{i=1}^{N} \mathcal{L}_{\text{D-SC}}^{(i)} + \beta \cdot \text{Tr}(\text{Cov}(Z)).
\end{equation}
This strategy mitigates the influence of hard negatives and enforces compact, informative representations for robust contrastive learning under sample heterogeneity.

\subsubsection{Gradient-Driven Adaptive Loss Scaling.}
This strategy addresses loss conflicts caused by varying sample difficulty in dynamic facial expression recognition. 
By leveraging a gradient-sensitivity-based adaptive scaling mechanism, it dynamically adjusts the weights of classification and contrastive losses to improve training stability and convergence.  
Inspired by SAM, we use the norm of adversarial gradient perturbations to estimate batch-level training difficulty. Let the gradient norm of the total loss be: $g = \|\nabla_w \mathcal{L}_{\text{total}}\|_2$, then compute the dynamic weights for each loss term as:
\begin{equation}
\lambda_{\text{CE}} = \frac{g}{g + 1} \cdot \alpha, \quad \lambda_{\text{SC}} = \frac{1}{g + 1} \cdot \beta ,
\end{equation}
where $\alpha$ and $\beta$ are base scaling coefficients. This strategy allows the model to emphasize classification loss during difficult training phases (large $g$) for stability, and focus on contrastive refinement during easier phases (small $g$).

\subsubsection{Overall Objective.}
The overall objective of \MB~ is defined as follows:
\begin{equation}
\mathcal{L}_{\text{total}} = \lambda_{\text{CE}} \cdot \mathcal{L}_{\text{CE}} + \lambda_{\text{SC}} \cdot \mathcal{L}_{\text{IB-D-SC}}.
\end{equation}
This optimization strategy enables per-batch awareness of training status and feature structure dynamics, achieving self-regulated balancing between classification and contrastive learning. As a result, \MB~significantly improves model robustness and generalization under real-world distribution shifts and individual variations.

\begin{table*}[t]
\renewcommand{\arraystretch}{0.9}
\centering
\setlength{\tabcolsep}{3mm}
\small

\scalebox{1.0}{
\begin{tabular}{ c|c|ccccccc|cc }
\hline
\multirow{2}{*}{Method} & \multirow{2}{*}{years} & \multicolumn{7}{c|}{Accuracy of Each Emotion(\%)} & \multicolumn{2}{c}{Metrics (\%)}  \\ 
\cmidrule(lr){3-11}
& & Hap.	  & Sad.    & Neu.   & Ang.	   & Sur.	& Dis.	  & Fea.	 & WAR   	& UAR \\ \hline
ResNet18+LSTM \cite{dfew} &ACM MM'20 &78.00 &40.65 &53.77 &56.83 &45.00 &4.14 &21.62 &53.08 &42.86 \\
EC-STFL \cite{dfew}  &ACM MM'20 &79.18 &49.05 &57.85 &60.98 &46.15 &2.76 &21.51 &56.51 &45.35 \\
Former-DFER \cite{formerDFER} &ACM MM'21   &84.05	 &62.57	   &67.52	&70.03	  &56.43	&3.45	 &31.78	    &65.70	   &53.69 \\ 
Logo-Former\cite{Ma2023LogoFormerLS} & ICASSP'23&85.39 &66.52 &68.94 &71.33 &54.59 &0.00 &32.71 &66.98  &54.21 \\
T-MEP \cite{10250883} &T-CSVT'24 &N/A&N/A&N/A&N/A&N/A&N/A&N/A&  68.85& 57.16
\\
CFAN-SDA \cite{Chen_Wen_Yang_Li_Chen_Wang} &T-CSVT'24  &\textbf{90.84} &\underline{70.91} &65.72 &69.97 &57.86 &\textbf{13.10} &\underline{35.36} &69.19& 57.70\\
GCA+IAL \cite{ial/aaai.v37i1.25077}   &AAAI'23         &87.95	 &67.21	   &\underline{70.10}	&\textbf{76.06}	  &\underline{62.22}	&0.00	 &26.44	    &69.24	   &55.71\\ 
M3DFEL \cite{m3dfel} &CVPR'23   &89.59 	 &68.38   &67.88	&\underline{74.24}	  &59.69	&0.00	 &31.63  	&69.25	   &56.10 \\  
RDFER\cite{10908623} & T-BIOM'25 &\underline{89.69} 	 &69.22  &\textbf{70.18}	&71.47	&62.08	&0.69	 &28.71  	&69.73	   &56.93  \\ 
LG-DSTF \cite{Zhang_Tian_Zhang_Guo_Xu} &T-MM'24&N/A&N/A&N/A&N/A&N/A&N/A&N/A& 69.82 & \underline{58.89}\\
CLIPER \cite{10687508}  &ICME'24 &N/A&N/A&N/A&N/A&N/A&N/A&N/A&  \underline{70.84}& 57.56 \\

\cmidrule(lr){1-11}
\name~(Ours)   &- &89.67 &\textbf{71.20}	&67.42	&73.03	&\textbf{64.44}	&\underline{12.41}	&\textbf{41.63}	&\textbf{71.60}	&\textbf{60.40} \\ \hline
\end{tabular}
}

\caption{Comparison (\%) of our Heterogeneity-aware Distributional Framework (\name) with state-of-the-art methods on DFEW 5-fd. (\textbf{Bold}: Best, \ul{Underline}: Second best)}

\label{res-dfew}
\vspace{-0.7cm}
\end{table*}

\begin{figure*}[t]
\centering
\includegraphics[width=0.9\textwidth]{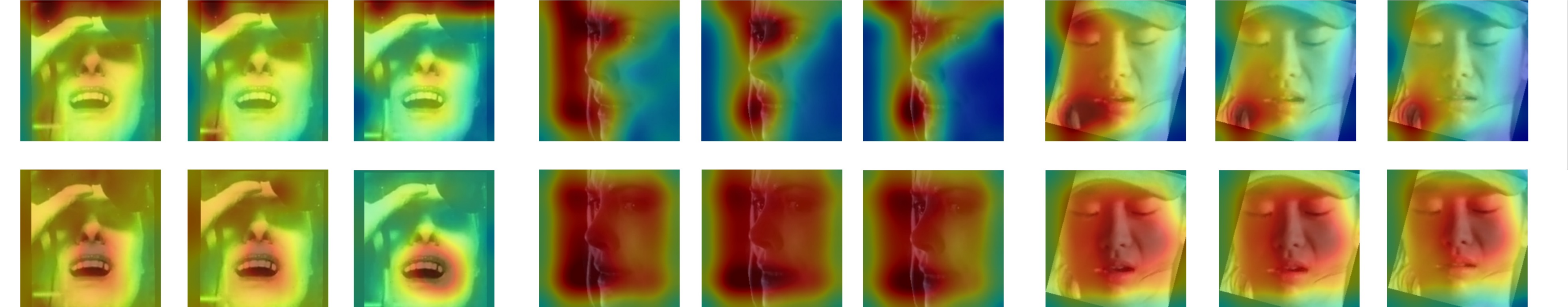} 
\caption{Visualization of the learned feature maps. There are three sequences are presented, which including the issues of the weak low illumination, occlusion, and non-frontal poses. For each sequence, the images in the first row are heatmaps generated by the baseline, and the images in the second row are heatmaps generated by our \name.}

\label{vis}
\vspace{-0.4cm}
\end{figure*}

\begin{figure*}[t]
\centering
\includegraphics[width=0.88\textwidth]{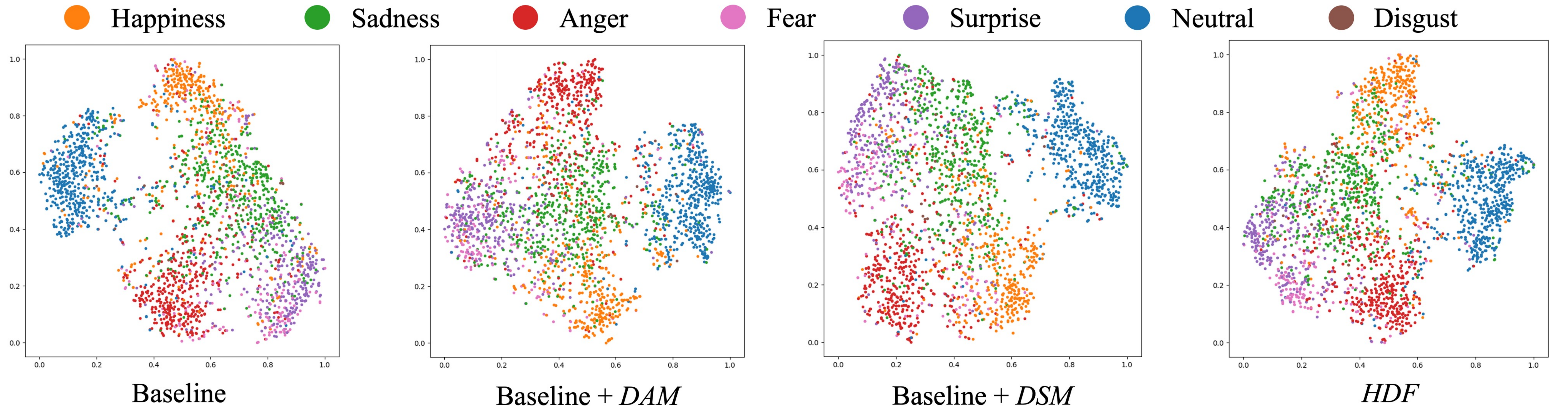} 
\caption{Illustration of feature distribution learned by the baseline and our proposed method in \name~on DFEW fd5.}
\vspace{-0.4cm}
\label{sne}
\end{figure*}

\begin{table}[h]
\renewcommand{\arraystretch}{1.0}
\centering
\setlength{\tabcolsep}{3mm}
\scalebox{0.9}{
\begin{tabular}{ c|cc }
\hline
\multirow{2}{*}{Method} & \multicolumn{2}{c}{Metrics (\%)}   \\ \cmidrule(lr){2-3}
  & WAR   & UAR \\ \hline
2ResNet18+LSTM \cite{ferv39k} &43.20 &31.28  \\
Former-DFER \cite{formerDFER} &46.85	&37.20	  \\
3D-DSwin \cite{10394270}    &46.98 &37.66  \\
GCA+IAL \cite{ial/aaai.v37i1.25077}     &48.54	&35.82	  \\
M3DFEL \cite{m3dfel} &47.67	&35.94	  \\ 
LG-DSTF \cite{Zhang_Tian_Zhang_Guo_Xu}  &48.19 &\underline{39.84}  \\ 
RDFER \cite{10908623}  &48.60 &36.47  \\ 
CFAN-SDA \cite{Chen_Wen_Yang_Li_Chen_Wang}&\underline{49.48} & 39.56\\

\cmidrule(lr){1-3} 
\name~(Ours)	&\textbf{50.30}		&\textbf{40.49}	\\ \hline
\end{tabular}
}
\caption{Comparison (\%) of our \name~with the state-of-the-art methods on FERV39K.}
\label{res-39k}
\vspace{-0.9cm}
\end{table}

\begin{table}[t]
\setlength{\tabcolsep}{1.4mm}
\renewcommand{\arraystretch}{1.0}

\centering

\scalebox{1.0}{
\begin{tabular}{c|cc|cc}
\hline
\multirow{2}{*}{Setting} & \multicolumn{2}{c|}{\name} & \multicolumn{2}{c}{Metric (\%)} \\ \cmidrule(lr){2-5} 
~ & \MA     & \MB & WAR & UAR           \\ \hline
a & \ding{55} & \ding{55} & 68.62 & 58.21\\
b & \CheckmarkBold    & \ding{55} &  69.73  & 59.02  \\  
c & \ding{55}  & \CheckmarkBold & 71.79    &  59.93    \\ 
d & \CheckmarkBold   & \CheckmarkBold    &\textbf{73.24} & \textbf{61.31}  \\ \hline
\end{tabular}}
\caption{Ablation study of different components in \name~on DFEW fd5. }

\label{ablation}
\vspace{-1.4cm}
\end{table}

\section{Experiments}
\subsection{Experimental Setup}
\subsubsection{Datasets}
We conduct extensive experiments on two popular in-the-wild DFER datasets, namely DFEW \cite{dfew} and FERV39k \cite{ferv39k}.

DFEW is a large-scale dynamic expressions in-the-wild dataset introduced in 2020, containing over 16000 video clips. These clips are collected from more than 1500 films worldwide and include various challenging disturbances such as extreme lighting and pose changes. Each clip is annotated by ten trained annotators under professional guidance and assigned to one of seven basic expressions: happy, sad, neutral, angry, surprised, disgusted, and fearful.

FERV39k is the largest available in-the-wild DFER dataset, containing 38935 video clips collected from 4 scenes, further subdivided into 22 fine-grained scenes. It is the first DFER dataset with large-scale 39K clips, scene-to-scene segmentation, and cross-domain support. Each video clip in FERV39k is annotated by 30 professional annotators to ensure high-quality labeling and assigned to one of the same seven main expressions as DFEW. We use the training and test sets provided by FERV39k for fair comparison.

\subsubsection{Metrics} 
In all experiments, to be consistent with previous methods, we choose weighted average recall (WAR) and unweighted average recall (UAR) as metrics. In reality, WAR and UAR hold different significance, \ie, WAR reflects the model’s overall effectiveness in real-world deployments, as it is influenced by major categories with a large number of samples. Since UAR can better reflect the model's performance in each category, more attention can be obtained by researchers on the model's balanced performance across various categories.

\subsubsection{Implementation Details} 
Our entire framework is implemented using PyTorch-GPU and trained on a single NVIDIA RTX A6000 GPU. In our experiment, all face images are resized to 160×160. We use augmentation techniques such as random cropping, horizontal flipping, and 0.4 color jitter. For each video, 16 frames are extracted as samples. The feature extraction network use the standard X3D model with pre-trained weights from Torchvision. The model is trained using AdamW as the base optimizer in combination with a cosine scheduler for a total of 100 epochs, including a 20-epoch warm-up period. The learning rate is set to 5e-4, with a minimum learning rate of 5e-6, and weight decay is set to 0.05. Based on prior work \cite{wu2023understanding}, the $\tau$ and $\eta$ in \MB~are set to 0.07 and 0.2 for stable and effective representation learning.

\subsection{Comparison with the State-of-the-art Methods}
 
\subsubsection{Results on DFEW}
Following previous works, we conduct experiments under a 5-fd cross-validation protocol. As shown in Tab.~\ref{res-dfew}, \name~achieves the best performance in both WAR and UAR metrics. In particular, it achieves consistently better performance on underrepresented and challenging categories, even surpassing some vision-language models, highlighting its strong generalization to heterogeneous and hard-to-learn samples. This demonstrates the effectiveness of our method in not only improving overall recognition accuracy, but also enhancing robustness under sample imbalance and expression diversity.

\subsubsection{Results on FERV39k}
As shown in Tab.~\ref{res-39k}, \name~achieves competitive results on the FERV39k, outperforming all baselines in both WAR and UAR. This confirms the effectiveness of our method in mitigating performance degradation caused by sample heterogeneity and distributional variations. Notably, we do not introduce any dataset-specific hyper-parameter tuning, applying the same training configuration to both DFEW and FERV39k. Despite the scale differences between the two datasets, our method consistently delivers strong results, highlighting its stability and ease of deployment in real-world scenarios without requiring extensive parameter adjustment.

\subsection{Ablation Studies}

\subsubsection{Effectiveness of Each Module}
We conduct ablation studies on DFEW fd5 to verify the individual contributions of our two core components in \name: the \MA~and the \MB. Results are shown in Tab.~\ref{ablation}.
Setting (a) employs only the X3D backbone with standard cross-entropy loss, serving as our baseline. Introducing \MA~alone (Setting b) brings noticeable improvements in both WAR and UAR, validating the effectiveness of distributionally-robust attention in suppressing noise and inconsistencies caused by hard samples. Adding \MB~alone (Setting c) leads to even higher gains, emphasizing the importance of dynamically balancing the optimization focus between classification and representation learning, particularly when hard examples dominate training dynamics. Finally, combining both modules (Setting d) enables \name~to achieve the best performance—outperforming the baseline by +4.62\% WAR and +3.10\% UAR. These results confirm the complementary nature of our modules in addressing sample heterogeneity from both the feature modeling and optimization perspectives.

\subsubsection{Effectiveness of Frequency and Temporal Branches in \MA}

We further conduct ablation studies to examine the individual contributions of the frequency and temporal branches within \MA~. As shown in Tab.~\ref{ablation-TFDRA}, enabling either the frequency branch (Setting b) or the temporal branch (Setting c) yields clear improvements over the baseline (Setting a), showing that each branch independently enhances robustness to a different type of hard sample: frequency attention mitigates performance drops caused by low-quality or stylistically biased samples, while temporal attention stabilizes learning under sequences with inconsistent dynamics. When both branches are jointly activated (Setting d), the model achieves the best overall performance, verifying that these components address complementary aspects of sample heterogeneity and jointly contribute to \MA’s robustness under real-world distribution shifts.

\subsection{Visualization.}

\subsubsection{Learned Feature Maps Visualization}
To verify the robustness of the proposed \name~under challenging conditions, we visualize the learned facial feature maps, as shown in Fig.~\ref{vis}, including cases with low illumination, occlusion, and non-frontal poses. In the first sequence, our method effectively suppresses the influence of occluded regions by focusing on informative visible areas, showcasing robustness against spatial ambiguity. In the second sequence, despite non-frontal views, our features still capture expressive regions, highlighting resilience to pose-induced heterogeneity. Furthermore, across frames with strong head motion, the model consistently attends to semantically stable regions, demonstrating robustness against temporal noise and maintaining reliable representation under complex variations.

\subsubsection{t-SNE Visualization}
We employ t-SNE \cite{SNE:v9:vandermaaten08a} to visualize the feature distributions learned by different model variants. As shown in Fig.~\ref{sne}, the baseline produces scattered and overlapping feature clusters, indicating limited discriminative capacity. Models enhanced with \MA~or \MB~yield noticeably tighter intra-category clusters and clearer inter-category boundaries, indicating stronger discriminability. When both modules are combined in the complete \name~framework, emotion categories become more distinctly separable. This confirms that our framework effectively mitigates representation ambiguity caused by sample heterogeneity, allowing the model to learn more robust and compact feature embeddings for DFER tasks.

\subsubsection{Visualization of Loss Landscape}
\begin{figure}[t]
\centering
\includegraphics[width=0.43\textwidth]{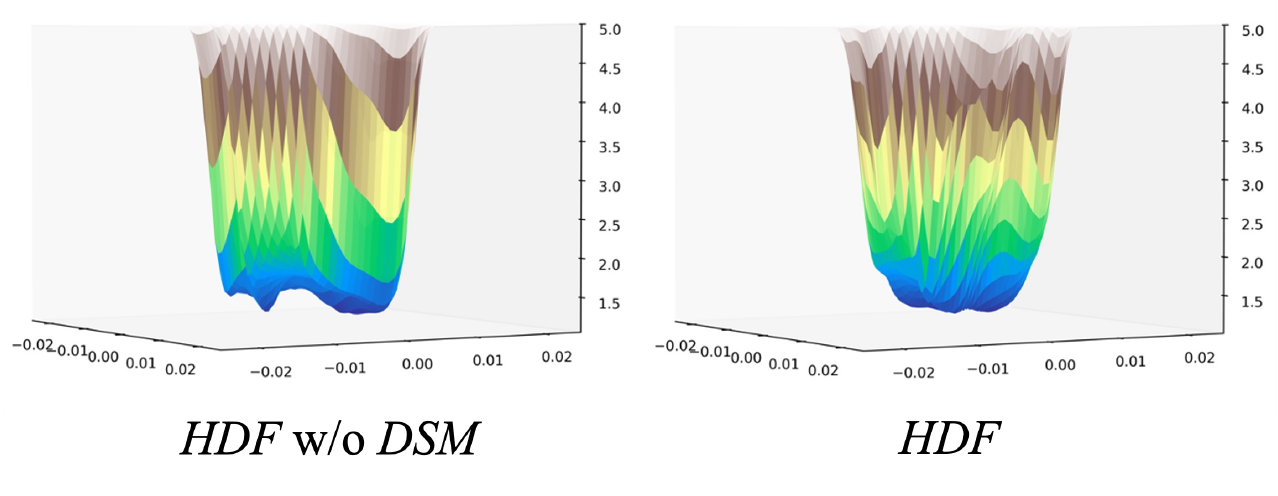} 
\caption{The visualization of separate loss landscape on DFEW fd5, optimized by AdamW (left) and our \MB~(right) respectively.}
\vspace{-0.6cm}
\label{land}
\end{figure}

To evaluate the effect of our \MB~on training dynamics, we visualize the loss landscape \cite{li2024improving}, as shown in Fig.~\ref{land}. Compared with the version without \MB~(left), our full \name~(right) exhibits a significantly flatter and smoother loss surface. This demonstrates that \MB~effectively suppresses gradient instability induced by hard or uncertain samples, guiding the model toward more stable optima. The improved smoothness highlights enhanced convergence stability and generalization capability, particularly in the presence of sample heterogeneity and hard samples caused by temporal ambiguity, style variation, or underrepresented categories in DFER.

\begin{table}[t]
\setlength{\tabcolsep}{1.4mm}
\renewcommand{\arraystretch}{0.85}

\centering

\scalebox{1.0}{
\begin{tabular}{c|cc|cc}
\hline
\multirow{2}{*}{Setting} & \multicolumn{2}{c|}{\MA} & \multicolumn{2}{c}{Metric (\%)} \\ \cmidrule(lr){2-5} 
 & Fre.     & Tim. & WAR & UAR           \\ \hline
a & \ding{55} & \ding{55} & 71.79  & 59.93\\
b & \CheckmarkBold    & \ding{55} & 72.60   & 60.43    \\  
c & \ding{55}  & \CheckmarkBold & 72.73   & 60.21    \\ 
d & \CheckmarkBold   & \CheckmarkBold    &\textbf{73.24} & \textbf{61.31}  \\ \hline
\end{tabular}
}
\caption{Ablation study of frequency and time branches in \MA~on DFEW fd5.  }

\label{ablation-TFDRA}
\vspace{-0.8cm}
\end{table}

\section{Conclusion}

We propose \name, a heterogeneity-aware optimization framework tailored for in-the-wild DFER, which explicitly addresses sample-level variations in style, dynamics, and category distribution. It integrates two key components: a dual-branch attention module, \MA, that performs time-frequency modeling to enhance robustness against visual inconsistency and temporal ambiguity, and a dynamic optimization strategy, \MB, which adaptively balances classification and contrastive objectives based on training dynamics to further mitigate the impact of hard or uncertain samples on representation learning. Together, these modules form a unified Distributionally Robust Optimization-based solution that effectively improves representation robustness and training stability under heterogeneous data conditions. Extensive experiments on DFEW and FERV39k demonstrate the superiority and generalizability of \name. In future work, we plan to extend this framework to multi-modal expression analysis and investigate deployment-friendly variants for real-world applications.
\section*{ACKNOWLEDGEMENTS}
This work is supported by Anhui Province Science Foundation for Youths (Grant No. 2308085QF230), Fundamental Research Funds for the Central Universities (Grant No. JZ2025HGTB0225), Major Scientific and Technological Project of Anhui Provincial Science and Technology Innovation Platform (Grant No. 202305a12020012), National Natural Science Foundation of China (Grant No. 62302145), and Students' Innovation and Entrepreneurship Foundation of University of Science and Technology of China (Grant No. CY2024X019A). We also thank the OpenI Community (https://openi.pcl.ac.cn) for its support.

\bibliographystyle{ACM-Reference-Format}
\bibliography{reference}
\end{CJK}
\end{document}